\documentclass[lettersize,journal]{IEEEtran}
\usepackage{amsmath,amsfonts}
\usepackage{algorithmic}
\usepackage{algorithm}
\usepackage{array}
\usepackage[caption=false,font=normalsize,labelfont=sf,textfont=sf]{subfig}
\usepackage{textcomp}
\usepackage{stfloats}
\usepackage{url}
\usepackage{verbatim}
\usepackage{graphicx}
\usepackage{cite}
\usepackage{graphicx}
\usepackage{amsmath}
\usepackage{amssymb}
\usepackage{booktabs}
\usepackage{multirow}
\usepackage{algorithm, algorithmic}
\usepackage{bbm}
\usepackage{color}
\usepackage{makecell}
\usepackage[switch]{lineno}

\usepackage[capitalize]{cleveref}
\crefname{section}{Sec.}{Secs.}
\Crefname{section}{Section}{Sections}
\Crefname{table}{Table}{Tables}
\Crefname{equation}{Eqn.}{Eqns.}
\crefname{table}{Tab.}{Tabs.}

\begin{document}

\title{Confidence Aware Learning for Reliable Face Anti-spoofing}

\author{Xingming Long,~\IEEEmembership{Student Member,~IEEE,},
        Jie Zhang,~\IEEEmembership{Member,~IEEE,},
        Shiguang Shan,~\IEEEmembership{Fellow,~IEEE,}
\thanks{This work is partially supported by Strategic Priority Research Program of the Chinese Academy of Sciences (No. XDB0680202), Beijing Nova Program (20230484368), Suzhou Frontier Technology Research Project (No. SYG202325), and Youth Innovation Promotion Association CAS.}
\thanks{This work involved human subjects or animals in its research. Approval of all ethical and experimental procedures and protocols was granted by Ethics Committee of Institute of Computing Technology, Chinese Academy of Sciences.}
\thanks{Xingming Long, Jie Zhang and Shiguang Shan are with the Key Laboratory of AI Safety of CAS, Institute of Computing Technology (ICT), Chinese Academy of Sciences (CAS), Beijing 100190, China, and also with the University of Chinese Academy of Sciences (UCAS), Beijing 100049, China (e-mail: xingming.long@vipl.ict.ac.cn; zhangjie@ict.ac.cn; sgshan@ict.ac.cn).}
\thanks{\textit{(Corresponding author: Jie Zhang)}}
}

\markboth{Journal of \LaTeX\ Class Files,~Vol.~14, No.~8, August~2021}%
{Shell \MakeLowercase{\textit{et al.}}: A Sample Article Using IEEEtran.cls for IEEE Journals}


\maketitle

\begin{abstract}
Current Face Anti-spoofing (FAS) models tend to make overly confident predictions even when encountering unfamiliar scenarios or unknown presentation attacks, which leads to serious potential risks. To solve this problem, we propose a Confidence Aware Face Anti-spoofing (CA-FAS) model, which is aware of its capability boundary, thus achieving reliable liveness detection within this boundary.
To enable the CA-FAS to ``know what it doesn't know'', we propose to estimate its confidence during the prediction of each sample. Specifically, we build Gaussian distributions for both the live faces and the known attacks. The prediction confidence for each sample is subsequently assessed using the Mahalanobis distance between the sample and the Gaussians for the ``known data''. We further introduce the Mahalanobis distance-based triplet mining to optimize the parameters of both the model and the constructed Gaussians as a whole.
Extensive experiments show that the proposed CA-FAS can effectively recognize samples with low prediction confidence and thus achieve much more reliable performance than other FAS models by filtering out samples that are beyond its reliable range.

\end{abstract}



\section{Introduction}
\label{sec:intro}
\IEEEPARstart{F}{ace} recognition models are widely applied in security fields such as access control and payment systems nowadays. However, the emergence of various presentation attacks, such as deceiving the system with a video or a printed photo, leads to serious security risks. To defend face recognition models from presentation attacks, face anti-spoofing (FAS) methods are proposed to serve as a protective measure. Existing FAS works have demonstrated excellent detection performance in intra-dataset settings \cite{LBP_HoG2012,Color2016,distortion2016,CNN2014,LSTM_CNN2015,RGBD2021,reflection2020}. However, when these FAS models encounter challenging cases from unseen scenarios (unseen domains) or unknown presentation attacks, they usually suffers from a noticeable performance decline. Some studies \cite{OCSCM_CVPR2024, STDN_PAMI2022} attempt to identify unknown attacks through one-class classification or by reconstructing live faces. However, these methods perform poorly when encountering live faces under various domains.

\begin{figure}[t]
  \centering
  \includegraphics[width=0.9\linewidth]{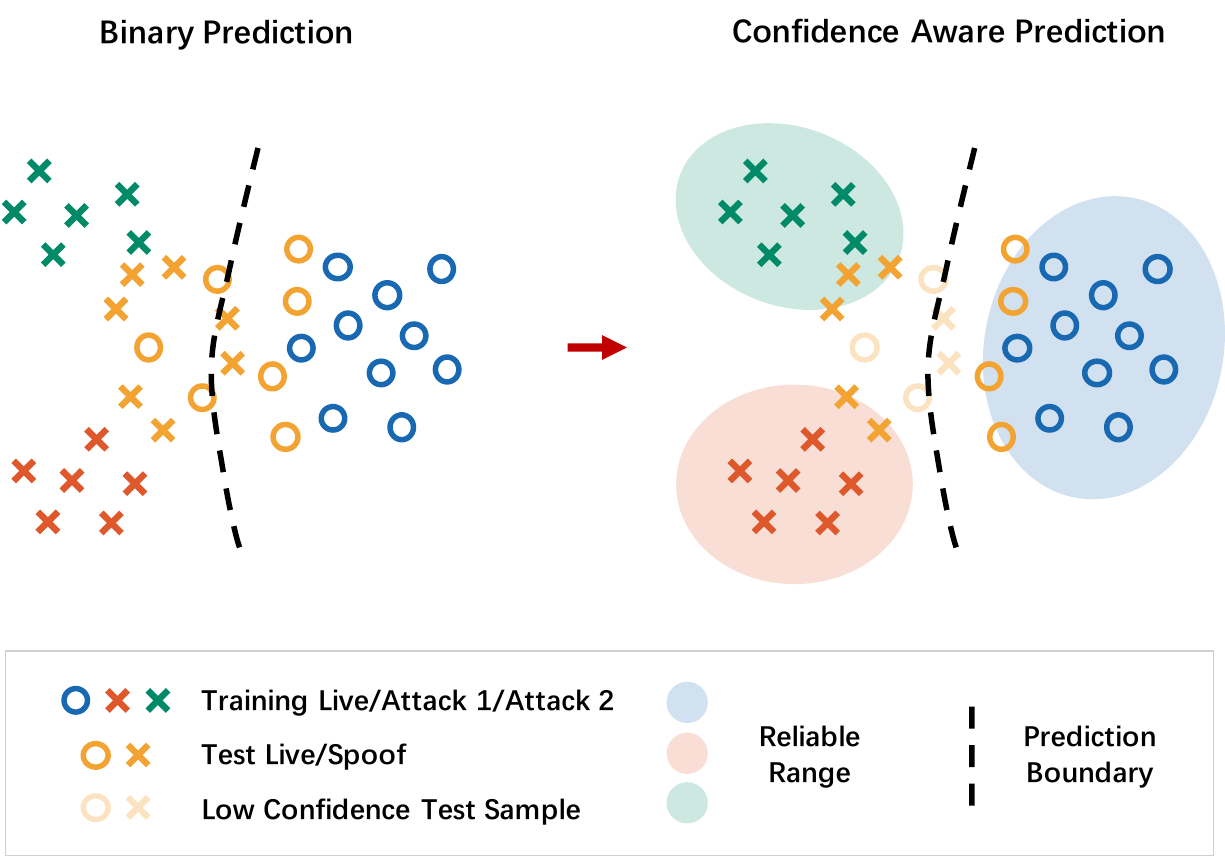}
  \caption{The relationship between the traditional binary prediction and the proposed confidence aware prediction. In binary prediction, the model uses a prediction boundary to divide the test data. However, due to unseen scenarios or unknown attacks, the differences between test and training feature distributions can lead to misclassifications, which poses serious risks. In contrast, the confidence aware method sets a confidence threshold to filter out samples beyond the reliable range, ensuring that only the samples with high prediction confidence are subjected to the classifier. The low confidence samples are rejected and passed for manual review, reducing the risk of misclassification.}
  \label{fig:intro}
\end{figure}

Recently, many works aim to enhance the cross-domain generalization of FAS models when the distribution of the target test dataset is inconsistent with the source training data. Domain adaptation (DA) methods are adopted to improve the model's performance on the target dataset by using a small amount of data from that dataset \cite{Unsupervised_DA2018, wang_improving_DA2019, Self_DA2021, DA_ViT_adapter, DA_trans_target_to_align, DA_source_free}. However, it is often impossible to know in advance what the unknown presentation attacks are in practical scenarios, making it challenging to obtain data from the target dataset for DA-based methods. Another category of works adopts the domain generalization (DG) methods, in which they simultaneously utilize multiple diverse training datasets to enhance the model's cross-domain generalization. A commonly used method of most DG-based FAS works is to align the feature space of different domains, i.e., various training datasets, for seeking domain invariant features\cite{MADDG_CVPR2019, DualReweighting2021, SSDG_CVPR2020, PatchNet_CVPR2022, SAFAS_CVPR2023}. Subsequent works resorts to feature disentanglement by removing biased factors in the training datasets, which enhances the model's generalization capability \cite{disentangle_attention2021, MixtureDomain_DDAM_AAAI2021, FaceID_CVPR2020, disentangle2022, SSAN_CVPR2022, DSCI_TIFS2023}. Additionally, meta-learning also serves as a crucial approach for achieving FAS domain generalization via simulating domain shifts in the training stage \cite{NAS_FAS2020, meta2020, fine_meta2020,Meta_Pattern_TIFS2022}. Although the DG-based methods can exploit the domain invariant feature from multiple datasets, they are still constrained by the limited types of presentation attacks in the training datasets. The models trained on these datasets always have a limited capability boundary and can hardly perform well on all the unknown presentation attacks. 

After analyzing the shortcomings of existing FAS methods, we find that the generalization ability of liveness detection models is inherently limited due to the constraints of the scenarios and attack types included in the training datasets. However, in this era of rapidly evolving attack techniques, new presentation attack methods will inevitably emerge that exceed the capabilities of current FAS models. As mentioned earlier, FAS models play a crucial role in safeguarding critical security domains where they are employed to protect face recognition systems. Any inaccuracies in the detection can result in serious consequences. Therefore, we propose the importance of reliable face anti-spoofing, which requires the FAS model to be aware of its own capability boundary. The model should operate within its capability boundary and reject the ``unknown samples'' outside this boundary. Only when the model ``knows what it doesn't know'' can security issues caused by overly confident misclassification be avoided.

In this work, we propose a Confidence Aware Face Anti-spoofing (CA-FAS) model, which can achieve reliable face anti-spoofing by measuring the prediction confidence of any input sample. 
Specifically, we build Gaussian distributions in the feature space for the live faces and each of the known presentation attacks, and realize the confidence aware prediction as shown in \Cref{fig:intro}. 
Each Gaussian represents a reliable range for the feature from a category, i.e., live or one of the known attacks. If an input's feature is close to a particular Gaussian, we consider the model's prediction for that sample to be reliable. Conversely, if a sample's feature is far from any known Gaussian distribution, we consider the model's prediction confidence for that sample to be low, and the model will refuse to make a prediction on it to avoid misclassification. We apply the Mahalanobis distance to measure the distance between an input feature and a constructed Gaussian, as it considers both the mean and covariance of the distribution and makes a more accurate assessment.

During training, we introduce a Mahalanobis distance-based triplet mining to optimize the parameters of both the FAS feature extractor and the constructed Gaussians. 
The triplet loss simultaneously pulls close the Mahalanobis distance between the feature of each category, i.e., live faces or each known attack, and their corresponding Gaussian, while pushing far the Mahalanobis distance between the Gaussians of different categories.
Finally, we design experiments with eight FAS datasets to evaluate the performance of FAS models on unseen scenarios and unknown presentation attacks. The results show that our CA-FAS model can effectively recognize samples beyond its capability boundary and achieves reliable liveness detection under different experimental settings.

The main contributions of this work are summarized as follows:

\begin{itemize}

\item We pull up the importance of reliable face anti-spoofing, which requires the FAS model to recognize its own capability boundary. This helps avoid making overly confident predictions on uncertain samples with unknown scenarios or presentation attacks, thereby mitigating potential security risks.

\item We propose the Confidence Aware Face Anti-spoofing (CA-FAS) model, which establishes its reliable range in the feature space and evaluates the prediction confidence through Mahalanobis distance. During inference, the proposed model can provide the prediction confidence, which reduces the risk of overly confident misclassification and achieves reliable face anti-spoofing.

\item We conduct experiments to evaluate the performance of FAS models when faced with unknown scenarios and presentation attacks. The results demonstrate the proposed CA-FAS achieves much more reliable performance by filtering out samples that are beyond its capability boundary.

\end{itemize}


\section{Related Work}
\label{sec:related}

\subsection{Conventional Face Anti-spoofing Methods}
Early face anti-spoofing methods are based on the vision hand-craft descriptors, using manually designed features to detect whether an image contains presentation attacks. Examples of such features used in liveness detection include LBP~\cite{LBP_HoG2012,LBP2015}, HoG~\cite{HoG2013}, SURF~\cite{Color2016}, and SIFT~\cite{distortion2016}. However, the performance of these methods is limited by the expressive power of hand-craft descriptors, and these methods often fall short in performance when faced with more complex attack scenarios. 
Therefore, researchers turn to using more powerful deep neural networks in subsequent FAS works, and great improvements are achieved in the field of FAS using convolutional neural networks (CNNs) ~\cite{CNN2014,Depth_CNN_2017} and long short-term memory (LSTM) ~\cite{LSTM_CNN2015}.
Due to neural networks' superior ability to integrate multimodal information, data from other modalities like depth maps~\cite{RGBD2021,Depth_CNN_2017} and reflection maps~\cite{reflection2019, reflection2020} are also utilized to enhance the performance of liveness detection models.
Despite the strong performance of the aforementioned methods in intra-dataset experiments, they are prone to overfitting the training data, leading to poor performance in unseen scenarios.

\subsection{FAS for Unknown Attacks}
Due to the continuous emergence of unknown presentation attacks, some studies explore how to enable models to detect novel attacks that are not included in the training data. A work resorts to the one-class classification method \cite{OCSCM_CVPR2024}, aiming to train the model only with live faces and generate an all-zero output for the live input (indicating that all positions are normal). When the model encounters a sample containing presentation attacks, the corresponding output will be activated at the positions of the attack, thereby producing a spoof cue map.
Other methods achieve the detection of unknown attacks by reconstructing the live faces \cite{despoofing_IJCB2023, STDN_PAMI2022}. The trained model can restore any input face to a live face, and the residual between the input and the live face can thus reflect the attack regions present in the input image.
However, these methods for detecting unknown attacks face a common problem: they struggle to effectively handle the data collected with different devices or environments, i.e., cross-domain data. Domain changes in live faces are easily perceived as anomalies by these methods, leading to misclassification.

\subsection{Cross-domain FAS methods}
Many efforts are devoted to enhancing the cross-domain capabilities of FAS models. One approach to enhance the cross-domain generalization of FAS models is through domain adaptation (DA) methods. Some of the researchers resort to adapter modules to fine-tune the model with data from the target dataset \cite{DA_ViT_adapter}. Others focus on stylizing the target data to match the source data \cite{DA_trans_target_to_align}. However, the limitation of the DA-based method lies in the requirement for a small amount of target data. In practical applications, obtaining samples with unknown potential presentation attacks in advance is not feasible.

Another commonly used approach to improve the FAS generalization is through domain generalization (DG) methods. The DG-based methods utilize multiple datasets simultaneously to enhance the generalization of FAS models. Most DG-based FAS works adopt the idea of feature alignment \cite{MADDG_CVPR2019,SSDG_CVPR2020,SAFAS_CVPR2023}, making the FAS feature indistinguishable across multiple training datasets. Subsequent works introduce the disentangle methods, selecting task-irrelevant features, such as identity features \cite{FaceID_CVPR2020, DSCI_TIFS2023} or style features \cite{SSAN_CVPR2022}, and disentangle them from the FAS feature to reduce the bias. Additionally, meta-learning is also an important research direction in FAS domain generalization. Different training datasets are used as meta-train sets and meta-test sets separately, thereby learning effective model structures \cite{NAS_FAS2020} and FAS patterns \cite{Meta_Pattern_TIFS2022}. A recent study encourages the model to converge towards an optimal flat minimum \cite{GACFAS_CVPR2024}, achieving great results.
\textcolor{black}{Another recent work further explores the use of the hierarchical relations in samples to enhance model's generalization ability across unseen domains \cite{hierarchical_CVPR2024}. Manipulation of the test-time data style is also considered a promising approach for boosting FAS performance \cite{Test-time-DG_CVPR2024}.}
However, due to the limitations of attack types in the training data, these DG-based methods still can never perform well on all the unknown presentation attacks. 


\section{Method}
\label{sec:method}

\subsection{Overview}

\begin{figure*}[t]
  \centering
  \includegraphics[width=0.8\linewidth]{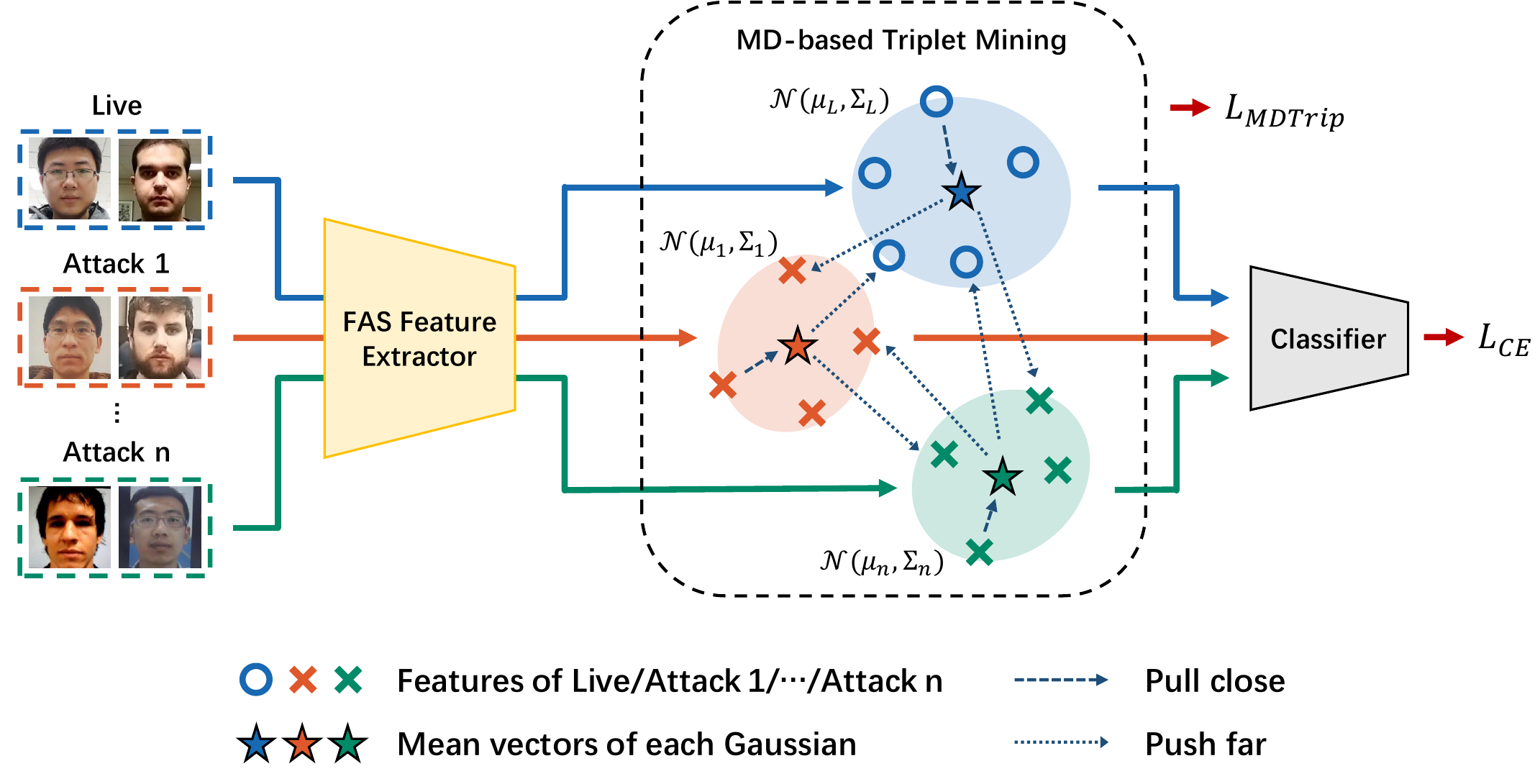}
  \caption{
  \textbf{The overall training process of the proposed CA-FAS.} In the feature space, we construct Gaussian distributions for each training category, i.e., live faces and each known attack, using learnable mean vectors and covariance matrices. We then apply Mahalanobis distance-based triplet optimization to minimize the distance of each sample to its corresponding category's Gaussian distribution and maximize its distance to other Gaussian distributions. Finally, the extracted features are fed into the liveness detection classifier, where the classification cross-entropy loss between the predicted results and the ground truth is calculated.}
  \label{fig:overview}
\end{figure*}

In this work, we propose the Confidence Aware FAS (CA-FAS) that is aware of its own capability boundary and thus achieves reliable liveness detection in the face of unseen scenarios or unknown presentation attacks. The overall training process of the framework is shown in \Cref{fig:overview}.
First, we construct Gaussian distributions in the feature space for live faces and each known attack type, using these distributions to evaluate the model's reliable range. Then, we optimize the parameters of the model and the Gaussians using the Mahalanobis distance-based (MD-based) triplet mining. The Mahalanobis distance-based triplet loss not only pulls the samples closer to their corresponding Gaussian but also pushes far the distance between the Gaussians of different categories, i.e., live and each known attack.
Once the model is trained, we can implement the confidence aware prediction. Specifically, by measuring the Mahalanobis distance of a sample to each Gaussian distribution, we use the negative of the minimum distance as the prediction confidence for the sample. We can then set an appropriate confidence threshold according to the requirements of the actual application, filtering out samples beyond the model's reliable range. The confidence aware prediction helps the model avoid security risks associated with misclassification of uncertain samples.

\subsection{Modeling of the Reliable Range}
To achieve reliable face anti-spoofing, we need the FAS model to understand the reliable range within its capability boundary so it can ``know what it knows''. We find that the model's reliable range is strongly related to the distribution of the training data: the more a test sample conforms to the training distribution, the more it falls within the model's reliable range, and the more accurate the model's prediction will be. Conversely, the model is more likely to make incorrect predictions on test samples that deviate from the training distribution.

In order to measure the reliable range of the proposed CA-FAS, we construct Gaussian distributions in the feature space to fit the training data. The distance between a feature and the training distribution is used to estimate the model's prediction confidence on the corresponding sample. To ensure that the extracted features are discriminative enough for the liveness detection, we build Gaussian distributions for the live faces and each known attack separately:
\begin{equation}
\begin{aligned}
f(X_L) &\sim \mathcal{N}(\mu_{L},\Sigma_{L}), \\
f(X_i) &\sim \mathcal{N}(\mu_{i},\Sigma_{i}), i\in \{1,2,...,n\},
\label{eq:Gaussian}
\end{aligned}
\end{equation}
where $X_L$ represents live faces, $X_i$ represents samples containing the $i^{th}$ known presentation attack, $n$ represents the number of known attacks, $f$ represents the feature extractor, $\mu_{L}$ and $\mu_{i}$ represent the mean vectors of each constructed Gaussian, $\Sigma_{L}$ and $\Sigma_{i}$ represents the covariance matrices.
Instead of using statistical measures, we employ trainable parameters as the mean vectors and covariance matrices of each Gaussian.

After modeling the training data using the Gaussians in the feature space, we can use the distance between a feature and these Gaussians to assess the model's prediction confidence on the corresponding input. The specific process of the confidence aware prediction will be introduced later.

\subsection{Mahalanobis Distance-based Triplet Mining}
The training objectives of the CA-FAS are twofold. First, to minimize the distance between the features and the corresponding Gaussian, ensure that the Gaussians accurately represent the distribution of the training data. Second, to maximize the distance between the features from different categories, i.e., live faces and each known attack, thereby enabling the features to be discriminative enough for liveness detection.

Based on the above considerations, we use a Mahalanobis distance-based triplet loss to optimize the parameters of the Gaussians as well as the feature extractor:
\begin{equation}
\begin{aligned}
L_{MDTrip} = &\sum_p \sum_{x\in X_p} \Big( MD\{f(x),\mathcal{N}(\mu_{p},\Sigma_{p})\} - \\
& \min_{q;q\ne p} MD\{f(x),\mathcal{N}(\mu_{q},\Sigma_{q})\} + margin \Big),
\label{eq:Triplet}
\end{aligned}
\end{equation}
where $p,q\in \{L\}\cup\{1,2,...,n\}$ represent a data category from either the live faces or one of the known presentation attacks, and $MD\{z,\mathcal{N}\}$ represents the Mahalanobis distance between a feature $z$ and a distribution $\mathcal{N}$ as follows:
\begin{align}
MD\{z,\mathcal{N}(\mu,\Sigma)\} = (z-\mu)^T\Sigma^{-1}(z-\mu).
\label{eq:Mahalanobis}
\end{align}
We apply Mahalanobis distance in the triplet mining as it considers both the mean and the covariance of the Gaussians, resulting in a more accurate assessment of the relationship between features and distributions.

Under the Mahalanobis distance-based triplet mining, features of samples from the same category are closely distributed within the corresponding Gaussian, while the distances between features from different categories are increased, as illustrated in \Cref{fig:overview}. Combined with the cross-entropy loss $L_{CE}$ used for training the feature-based FAS classifier, the final optimization objective of the whole CA-FAS framework is as follows:
\begin{equation}
  L_{total} = L_{CE} + \lambda L_{MDTrip},
  \label{eq:total loss}
\end{equation}
where $\lambda$ is the balanced parameter.

\subsection{Confidence Aware Prediction}
A brief schematic of the confidence aware prediction can be referenced in \Cref{fig:intro}. The model will reject samples that fall beyond the reliable range, and such samples should be referred to manual review to avoid security risks caused by misclassification.
Specifically, for each input sample, we calculate the Mahalanobis distances between its feature and each of the constructed Gaussian distributions. We then take the minimum of these distances as the sample's deviation from the reliable range, and its negative is used as the model's prediction confidence. The calculation is as follows:
\begin{align}
\mathcal{C}(x) = &- \min_p MD\{f(x),\mathcal{N}(\mu_{p},\Sigma_{p})\}, \\
&p\in \{L\}\cup\{1,2,...,n\}.
\label{eq:conf_score}
\end{align}

During the prediction process, a confidence threshold $\theta_c$ is set for the CA-FAS model. When the confidence score of an input exceeds the threshold, the model will conduct the liveness detection via the classifier. If the confidence score of the input is below the threshold, the model will reject the sample and provide an alert to the user, indicating the possibility of unknown scenarios or presentation attacks. The output of the confidence aware prediction is as follows:
\begin{align}
Output = \left\{
\begin{array}{ll}
g(f(x)) & \text{if } \mathcal{C}(x) \ge \theta_c \\
\text{alert} & \text{if } \mathcal{C}(x) < \theta_c
\end{array}
\right.,
\label{eq:conf_thr}
\end{align}
where $g$ represents the feature-based FAS classifier.

The confidence threshold $\theta_c$ can be set as an empirical constant value or based on a specific quantile $p$ of the prediction confidence across the entire test data $X_{test}$:
\begin{align}
\theta_c(p) = \underset{\theta}{\arg} \{P(\mathcal{C}(x) < \theta)=p| x\in X_{test}\}.
\label{eq:conf_quant}
\end{align}

In subsequent experiments, because some test datasets deviate significantly from the training data, to avoid all data being classified as anomalous after applying a constant confidence threshold, we apply the quantile threshold.


\section{Experiments}
\label{sec:experiments}

\subsection{Experimental Settings}

\textbf{Datasets and Protocol.} We take four public datasets as the training data in our experiments: MSU-MFSD \cite{MSU2015}, CASIA-FASD \cite{CASIA2012}, Idiap Replay-Attack \cite{REPLAY2012}, and OULU-NPU \cite{OULU2017}. All four datasets above include the two most common presentation attacks: print attacks and video replay attacks.
We employ several other typical FAS datasets as the test data for the experiments: HKBU-MARs \cite{HKBU2016}, ROSE-Youtu \cite{Unsupervised_DA_Rose-Youtu2018}, SiW \cite{SiW2018} and WMCA \cite{WMCA2019}. Each of these datasets contains different scenarios or presentation attacks, which differ from the four training datasets to varying degrees.
Regarding the types of presentation attacks contained in each dataset, HKBU-MARs only includes spoof data with 3D mask, and SiW only includes spoof data with print and replay attacks. ROSE-Youtu and WMCA each include different types of mask attacks in addition to print and replay attacks. Additionally, these test datasets' domain information is different from the training data due to the various devices used, lighting conditions, background settings, and other factors during data collection. For example, part of the data in WMCA is collected with a uniform green background, which is significantly different from the training data.
All data undergo face alignment preprocessing using RetinaFace~\cite{RetinaFace2019}. It is noteworthy that WMCA also contains makeup attacks; however, these data are not recognized as faces in our face alignment preprocessing. Therefore, results for makeup attacks are not involved in subsequent experiments.

\textbf{Implementation Details.} 
We use the ResNet-18 \cite{ResNet2016} as the backbone for the feature extractor of CA-FAS. The feature dimension of the model is set to 512, and thus the Gaussian distributions in the model also have a dimension of 512. We consider all the print attacks from the four training datasets as one attack type and the replay attacks as the other attack type. Therefore, our model includes parameters for a total of three Gaussian distributions to be trained, i.e., print attacks, replay attacks, and live faces. The momentum for SGD is set to 0.9, and the weight decay is set to 5e-4.

\subsection{Main Results}

\begin{table*}[t]
\centering
\caption{Comparison Results on Test Datasets with Unseen Scenarios and Unknown Attacks.}
\resizebox{0.85\hsize}{!}{
\begin{tabular}{ccccccccccc}
        \toprule
         & \multicolumn{2}{c}{\textbf{HKBU-MARs}} & \multicolumn{2}{c}{\textbf{ROSE-Youtu}} & \multicolumn{2}{c}{\textbf{SiW}} & \multicolumn{2}{c}{\textbf{WMCA}} & \multicolumn{2}{c}{\textbf{Mean}} \\
         \cmidrule(r){2-3}\cmidrule(r){4-5}\cmidrule(r){6-7}\cmidrule(r){8-9}\cmidrule(r){10-11}
         & HTER               & AUC               & HTER            & AUC             & HTER            & AUC             & HTER            & AUC             & HTER            & AUC             \\
         \midrule
Baseline                       & 22.79                    & 85.29                   & 13.93                    & 93.52                   & 8.10                     & 97.62                   & 23.40                    & 85.56                   & 17.06                    & 90.50                   \\
SSDG \cite{SSDG_CVPR2020}      & 25.37                    & 82.76                   & 14.56                    & 93.00                   & 7.52                     & 96.43                   & 19.02                    & 88.31                   & 16.62                    & 90.12                   \\
SSAN \cite{SSAN_CVPR2022}      & 29.88                    & 78.27                   & 17.51                    & 90.48                   & 15.87                    & 91.89                   & 27.91                    & 79.46                   & 22.79                    & 85.02                   \\
SA-FAS \cite{SAFAS_CVPR2023}   & 31.07                    & 75.69                   & 23.15                    & 84.19                   & 6.44                     & 98.29                   & 17.07                    & 91.68                   & 19.43                    & 87.46                   \\
GAC-FAS \cite{GACFAS_CVPR2024} & 33.73                    & 71.56                   & 27.20                    & 80.23                   & 7.46                     & 97.74                   & 16.78                    & 90.91                   & 21.29                    & 85.11                   \\
         \midrule
OC-SCM \cite{OCSCM_CVPR2024}   & 52.55                    & 44.65                   & 41.32                    & 63.20                   & 54.57                    & 46.06                   & 39.21                    & 59.82                   & 46.91                    & 53.43                   \\
STDN \cite{STDN_PAMI2022}      & 33.95                    & 71.72                   & 25.13                    & 82.45                   & 8.07                     & 97.54                   & 31.21                    & 72.25                   & 24.59                    & 80.99                   \\
         \midrule
CA-FAS (p=0.0)                 & 23.82                    & 83.94                   & 14.88                    & 91.37                   & 7.77                     & 97.57                   & 17.96                    & 90.34                   & 16.11                    & 90.81                   \\
CA-FAS (p=0.3)                 & 18.66                    & 87.95                   & 8.94                     & 94.64                   & 6.43                     & 98.38                   & 16.92                    & 92.68                   & 12.74                    & 93.41                   \\
CA-FAS (p=0.5)                 & 15.60                    & 90.90                   & 3.91                     & 98.10                   & 2.61                     & 99.38                   & 13.13                    & 94.94                   & 8.81                     & 95.83                   \\
CA-FAS (p=0.7)                 & \textbf{14.52}           & \textbf{93.13}          & \textbf{1.37}            & \textbf{99.10}          & \textbf{1.12}            & \textbf{99.75}          & \textbf{4.71}            & \textbf{98.35}          & \textbf{5.43}            & \textbf{97.58}          \\
\bottomrule
\end{tabular}
}
\label{tab:main}
\end{table*}

The main experimental results are shown in \Cref{tab:main}. At the top of the table, ``Baseline'' represents the results of using only the ResNet-18 model for the binary classification of live and spoof faces. Other entries list the results of the state-of-the-art (SOTA) models using domain generalization (DG) methods.
We find that the most recent DG-based works, such as SA-FAS and GAC-FAS, exhibit unstable performance across different test datasets. They perform better than ``Baseline'' on certain datasets like SiW and WMCA but show decreased performance on other datasets such as HKBU-MARs and ROSE-Youtu.
Upon further analysis, we find that these methods perform best on the SiW dataset, which only includes spoof data with print and replay attacks, while they perform worst on the HKBU-MARs dataset, which contains 3D mask attacks that are not included in the training data.
This indicates that these DG-based methods do not seem to handle unknown attacks well.

In the middle of the table, OC-SCM and STDN are two methods designed to handle unknown attacks. However, it can be observed that these methods actually struggle to detect unknown attacks as all the test datasets contain domain changes.

From the above results, it can be concluded that the capability boundary of FAS models is limited under certain training data. Although some DG-based methods can improve their performance in unseen domains (like SiW), they do so at the expense of performance on unknown attacks (like HKBU-MARs).
If we do not assess a FAS model's reliability for the test environment, the poor performance on the data beyond the model's capability boundary can pose significant security risks for practical liveness detection applications.

At the bottom of the table are the results of the proposed CA-FAS model. It shows the results of the confidence aware prediction after filtering out low confidence samples at different quantile thresholds $p$ defined in \Cref{eq:conf_quant}.
When our method is evaluated on the entire test dataset ($p=0.0$), its performance is superior to ``Baseline'' on SiW and WMCA but slightly declines on HKBU-MARs and ROSE-Youtu. However, after filtering out samples with low prediction confidence, our method shows a significant improvement.
When setting the quantile threshold $p=0.3$, which removes the lowest $30\%$ of samples in terms of prediction confidence, the proposed CA-FAS achieves the best performance across all test datasets. As the confidence threshold continues to increase, the predictive performance of CA-FAS further improves. This demonstrates that our method clearly understands its own capability boundary, knowing which data it can predict more accurately and which data is beyond its capacity.
This understanding of the capability boundary is crucial for the reliability of algorithms in unknown detection environments. CA-FAS does not blindly make erroneous predictions when encountering samples beyond its reliable range, effectively reducing the risk associated with misclassification.

\subsection{Effectiveness of the MD-based Confidence}

\begin{table*}[t]
\centering
\caption{Comparison Between Different Implementation of Prediction Confidence.}
\resizebox{0.85\hsize}{!}{
\begin{tabular}{ccccccccccc}
        \toprule
         & \multicolumn{2}{c}{\textbf{HKBU-MARs}} & \multicolumn{2}{c}{\textbf{ROSE-Youtu}} & \multicolumn{2}{c}{\textbf{SiW}} & \multicolumn{2}{c}{\textbf{WMCA}} & \multicolumn{2}{c}{\textbf{Mean}} \\
         \cmidrule(r){2-3}\cmidrule(r){4-5}\cmidrule(r){6-7}\cmidrule(r){8-9}\cmidrule(r){10-11}
         & HTER               & AUC               & HTER            & AUC             & HTER            & AUC             & HTER            & AUC             & HTER            & AUC             \\
         \midrule
Baseline (p=0.0) & 22.79                    & 85.29                   & 13.93                    & 93.52                   & 8.10                     & 97.62                   & 23.40                    & 85.56                   & 17.06                    & 90.50                   \\
Baseline (p=0.3) & 16.46                    & 90.13                   & 13.96                    & 94.51                   & 12.86                    & 95.22                   & 24.15                    & 84.66                   & 16.86                    & 91.13                   \\
Baseline (p=0.5) & 12.51                    & 93.20                   & 12.34                    & 95.56                   & 10.90                    & 96.02                   & 25.65                    & 85.25                   & 15.35                    & 92.51                   \\
Baseline (p=0.7) & 9.80                     & \textbf{94.93}          & 7.55                     & 97.57                   & 6.54                     & 97.30                   & 10.49                    & 97.59                   & 8.59                     & 96.85                   \\
         \midrule
SSDG (p=0.0)     & 25.37                    & 82.76                   & 14.56                    & 93.00                   & 7.52                     & 96.43                   & 19.02                    & 88.31                   & 16.62                    & 90.12                   \\
SSDG (p=0.3)     & 20.34                    & 87.34                   & 9.36                     & 95.94                   & 8.92                     & 94.72                   & 31.78                    & 75.49                   & 17.60                    & 88.37                   \\
SSDG (p=0.5)     & 15.60                    & 90.60                   & 6.14                     & 97.21                   & 7.07                     & 95.49                   & 40.77                    & 68.82                   & 17.39                    & 88.03                   \\
SSDG (p=0.7)     & \textbf{9.32}            & 93.62                   & 4.33                     & 98.13                   & 5.17                     & 96.81                   & 38.15                    & 71.99                   & 14.24                    & 90.14                   \\
         \midrule
CA-FAS (p=0.0)   & 23.82                    & 83.94                   & 14.88                    & 91.37                   & 7.77                     & 97.57                   & 17.96                    & 90.34                   & 16.11                    & 90.81                   \\
CA-FAS (p=0.3)   & 18.66                    & 87.95                   & 8.94                     & 94.64                   & 6.43                     & 98.38                   & 16.92                    & 92.68                   & 12.74                    & 93.41                   \\
CA-FAS (p=0.5)   & 15.60                    & 90.90                   & 3.91                     & 98.10                   & 2.61                     & 99.38                   & 13.13                    & 94.94                   & 8.81                     & 95.83                   \\
CA-FAS (p=0.7)   & 14.52                    & 93.13                   & \textbf{1.37}            & \textbf{99.10}          & \textbf{1.12}            & \textbf{99.75}          & \textbf{4.71}            & \textbf{98.35}          & \textbf{5.43}            & \textbf{97.58}     \\
\bottomrule
\end{tabular}
}
\label{tab:eff_conf}
\end{table*}

\textbf{Comparative Results with Different Thresholds.}
We conduct comparative experiments to analyze the effectiveness of our prediction confidence based on Mahalanobis distance.
Since existing FAS models cannot directly evaluate their prediction confidence, we add a simple confidence calculation method based solely on the output for the experiments. 
Specifically, we assume that the closer the softmax probability of a classifier is to 1, the higher the confidence in its prediction. Therefore, we use the higher of the two softmax outputs from the binary classifier of the existing FAS model as its prediction confidence.
As SSDG obtains the overall best performance within the DG-based methods in \Cref{tab:main}, we select SSDG together with ``Baseline'' and add the output-based prediction confidence for the comparative experiments.
The results of the two models after filtering out low confidence samples at different quantile thresholds $p$ are shown in \Cref{tab:eff_conf}.

The results indicate that the proposed output-based prediction confidence also helps the models understand their capability boundaries to some extent. Both ``Baseline'' and SSDG achieve performance improvements on HKBU-MARs and ROSE-Youtu by filtering out uncertain samples.
\textcolor{black}{The ``Baseline'' with the output-based confidence even outperforms the proposed CA-FAS on HKBU-MARs across different values of $p$. This is because the model's output tends to be label-overfitted under the cross-entropy optimization, making the output-based confidence more effective in evaluating the uncertainty introduced by the unknown 3D mask attacks in HKBU-MARs.
However, we also observe that ``Baseline'' and SSDG do not show consistent improvement with increasing thresholds on SiW and WMCA, where unknown domains play a more significant role in making the model's predictions unreliable. This suggests that the output-based confidence is not stable in all the test conditions.}

In contrast, the Mahalanobis distance-based confidence in CA-FAS provides greater stability in unknown scenarios and attacks, consistently enhancing the model's prediction reliability across all four test datasets by filtering out low-confidence samples.


\begin{figure*}[t]
  \centering
  \includegraphics[width=\linewidth]{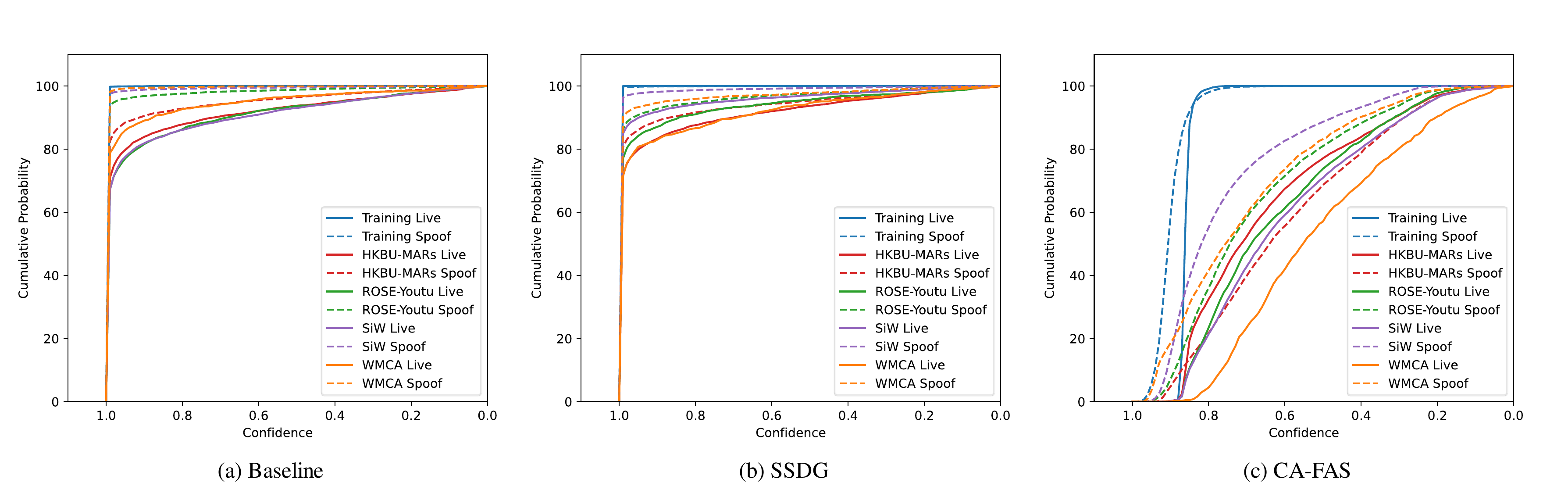}
  \caption{
  The Cumulative Distribution Function (CDF) graph of the prediction confidence from Baseline, SSDG, and CA-FAS. A higher curve indicates overall higher prediction confidence, while a lower curve indicates overall lower prediction confidence.}
  \label{fig:CDF}
\end{figure*}

\textbf{Distribution of the Prediction Confidence.}
We investigate the distribution of prediction confidence across different datasets to further examine the relationship between the model's prediction confidence and its performance. We use the output-based confidence obtained from ``Baseline'' and SSDG as comparisons. After normalization, the distributions of the confidence are presented using the Cumulative Distribution Function (CDF) graph, as shown in \Cref{fig:CDF}.

It can be observed that the CDF curves for ``Baseline'' and SSDG are overly concentrated in the upper left corner of the plot, indicating that for most samples, the models' confidence is close to 1. Overly confident predictions make it hard for the model to distinguish between samples within and beyond its capability boundary. This explains why the performance of ``Baseline'' and SSDG is unstable in \Cref{tab:eff_conf}.

Additionally, in practical applications, since it's challenging to obtain the quantile threshold in \Cref{eq:conf_quant} by pre-measuring the prediction confidence on all the test data, setting a fixed threshold is more feasible. 
However, the concentrated confidence distribution from ``Baseline'' and SSDG shown in \Cref{fig:CDF} would make it hard to select an appropriate fixed threshold.

For the proposed CA-FAS, since the domain information and attack types of the test datasets differ from the training data, the overall prediction confidence on test datasets is lower compared to the training dataset, as expected.
On datasets like SiW, which contains attack types consistent with the training data, the model exhibits relatively higher prediction confidence, leading to better performance. Conversely, for datasets like HKBU-MARs, which contains entirely different attack types, and WMCA, which shows significant background variations, the model's prediction confidence is lower, resulting in relatively poorer performance on these datasets. This further demonstrates the CA-FAS method's awareness of its own capability boundary, which effectively helps the model avoid making erroneous predictions on uncertain samples.
Besides, the confidence distribution of CA-FAS is more reasonable compared with ``Baseline'' and SSDG, facilitating the setting of appropriate thresholds to filter out low confidence samples in practical applications.

Lastly, an interesting phenomenon is observed: regardless of the prediction confidence calculation method, live faces have lower prediction confidence than spoof faces on certain datasets, such as WMCA. This contradicts the common assumption that the characteristics of live faces from different domains are consistent. The result indicates that significant domain shifts, such as the notable background changes in WMCA, might have great impacts on the model's detection on live faces.


\begin{figure*}[t]
  \centering
  \includegraphics[width=0.9\linewidth]{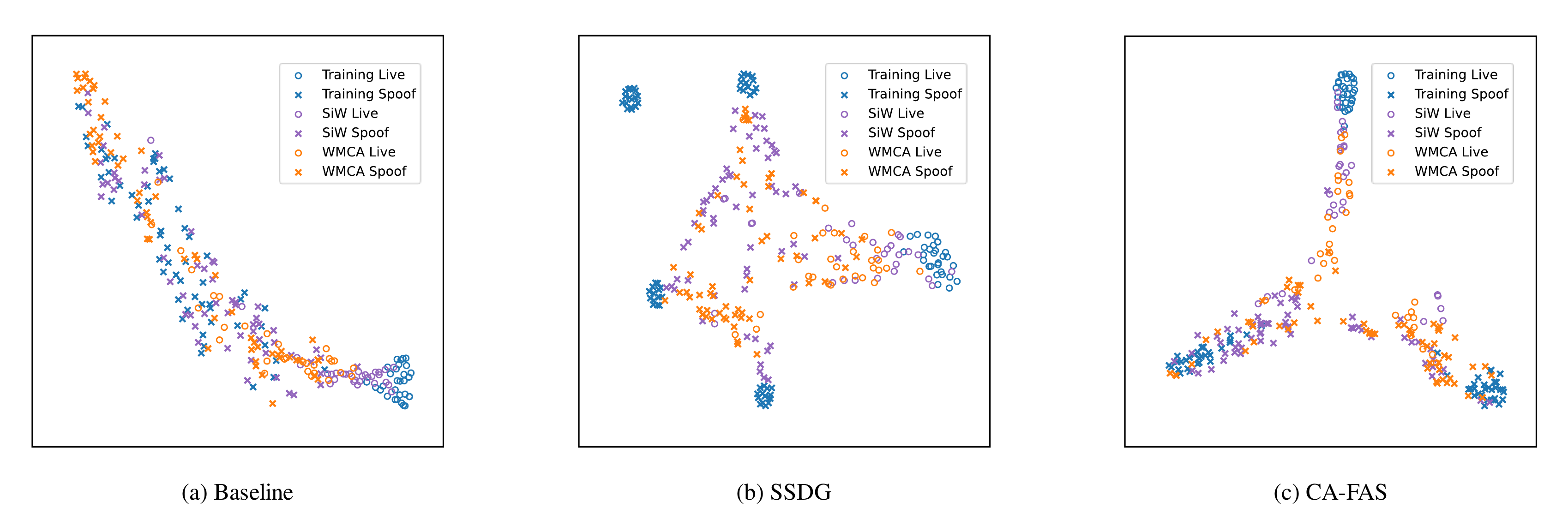}
  \caption{
  The t-SNE~\cite{tSNE2008} visualizations of the FAS features extracted by Baseline, SSDG, and CA-FAS from the training datasets and two target datasets (SiW and WMCA).}
  \label{fig:tSNE}
\end{figure*}

\textbf{Visualization of the Features.}
We visualize the feature distributions of above methods, i.e., Baseline, SSDG, and CA-FAS, using t-SNE~\cite{tSNE2008}, as shown in \Cref{fig:tSNE}. Specifically, we randomly sample 50 live faces from the training datasets, i.e., a mixed dataset containing MSU-MFSD, CASIA-FASD, Idiap Replay-Attack, and OULU-NPU, and 50 spoof faces from each of the two attack types, resulting in a total of 150 training samples for the feature visualization. For each test dataset, we randomly sample 50 live faces and 100 spoof faces for the visualization. To ensure the clarity and readability of the figure, we only draw the features of two test datasets, SiW and WMCA, which show the highest and lowest average prediction confidence in \Cref{fig:CDF}, respectively.

In ``Baseline'', due to the lack of constraints on the features, all features are approximately linearly distributed. Although the live and spoof features of the training data are well separated, many test live features are also distributed on the side of the training spoof features, leading to potential misclassifications. In SSDG, which constrains the training spoof features by domain, spoof features are clustered at four distinct locations corresponding to the four training domains. However, since the test data often do not share the same domain information as the training data, the features of the test spoof faces do not closely align with any of the training domains, impacting the model's ability to classify these samples.

In the proposed CA-FAS, the training features are well distributed near the preset three Gaussian distributions, i.e., print attacks, replay attacks, and live faces. For the test samples, those with features similar to the training data are distributed closer to the centers of these three Gaussians, indicating higher prediction confidence. We observe that the closer the features are to the centers of the three Gaussians, the more consistent the sample categories, i.e., live or spoof, are with the corresponding Gaussian, indicating better prediction performance for the model on higher confidence samples.
We also notice that some test live and spoof features are mixed in certain regions, indicating that the model struggles to make accurate predictions for these samples. However, these mixed regions are often far from the three known Gaussians, suggesting that the model has lower prediction confidence for these samples. This demonstrates the capability of CA-FAS to recognize the samples beyond its reliable range, thereby avoiding misclassification.

\subsection{Analysis of Samples with Different Confidence}

\begin{figure*}[t]
  \centering
  \includegraphics[width=\linewidth]{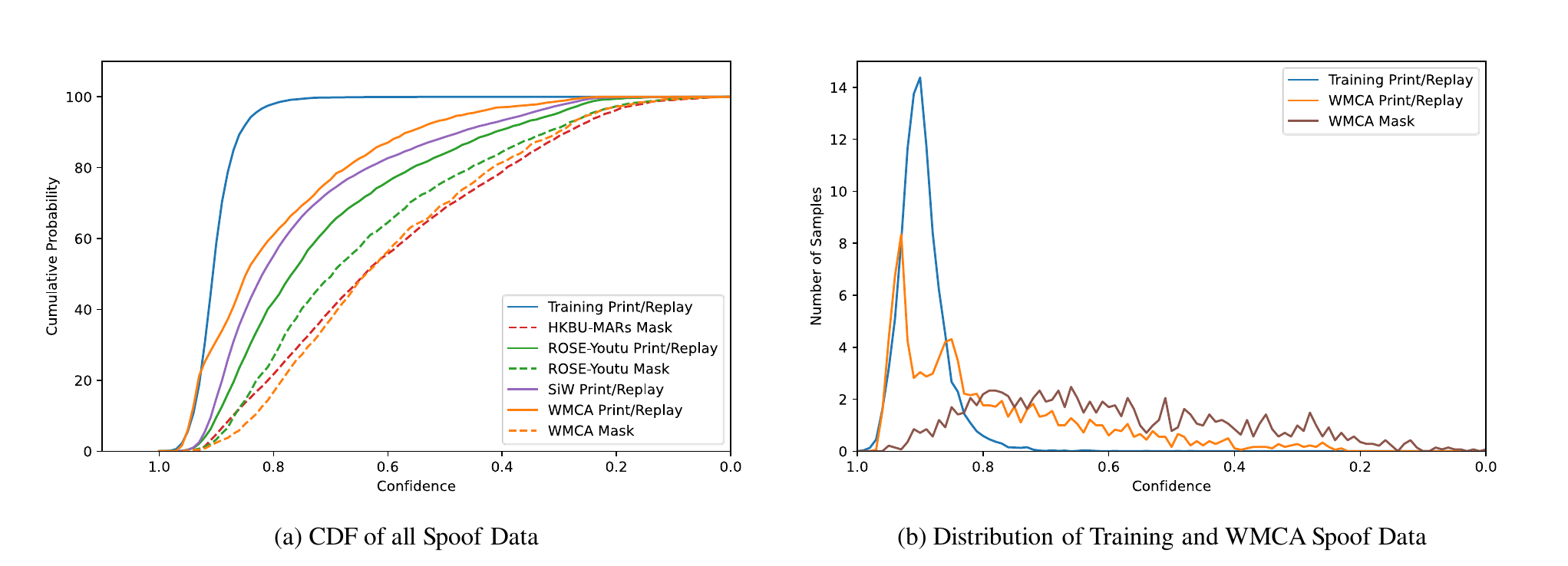}
  \caption{
  CA-FAS model's prediction confidence distribution for different types of spoof data. It can be seen that the overall prediction confidence for the mask attacks, which are not included in the training data, is lower than the confidence for print and replay attacks.}
  \label{fig:Fake_CDF}
\end{figure*}

We further compare the prediction confidence distribution of CA-FAS for different presentation attacks, as shown in \Cref{fig:Fake_CDF}. It can be observed that the model's prediction confidence for known attack types, i.e., print attacks and replay attacks, is generally higher than for the mask attacks that are not included in the training data. This indicates that the proposed CA-FAS can recognize the presence of unknown attacks in the test data. The model assigns lower prediction confidence to these samples with unknown attacks, suggesting that such samples should be flagged for manual review to avoid the security risks associated with misclassification.
We also notice that the model's prediction confidence for mask attacks in ROSE-Youtu is slightly higher than for mask attacks in HKBU-MARs and WMCA. This is because the mask attacks in ROSE-Youtu are all ``paper masks'', which are created by cutting out a printed face. In a sense, these paper masks can also be seen as a type of print attack, which explains why the model has relatively higher confidence in its predictions for these samples.

To eliminate the influence of the domain information from different test datasets, we specifically examine the prediction confidence distribution for the WMCA spoof faces and plot it on the right. It is more apparent from the figure that the model's prediction confidence for mask attacks in WMCA is generally lower than its confidence for print and replay attacks.

\begin{figure}[t]
  \centering
  \includegraphics[width=\linewidth]{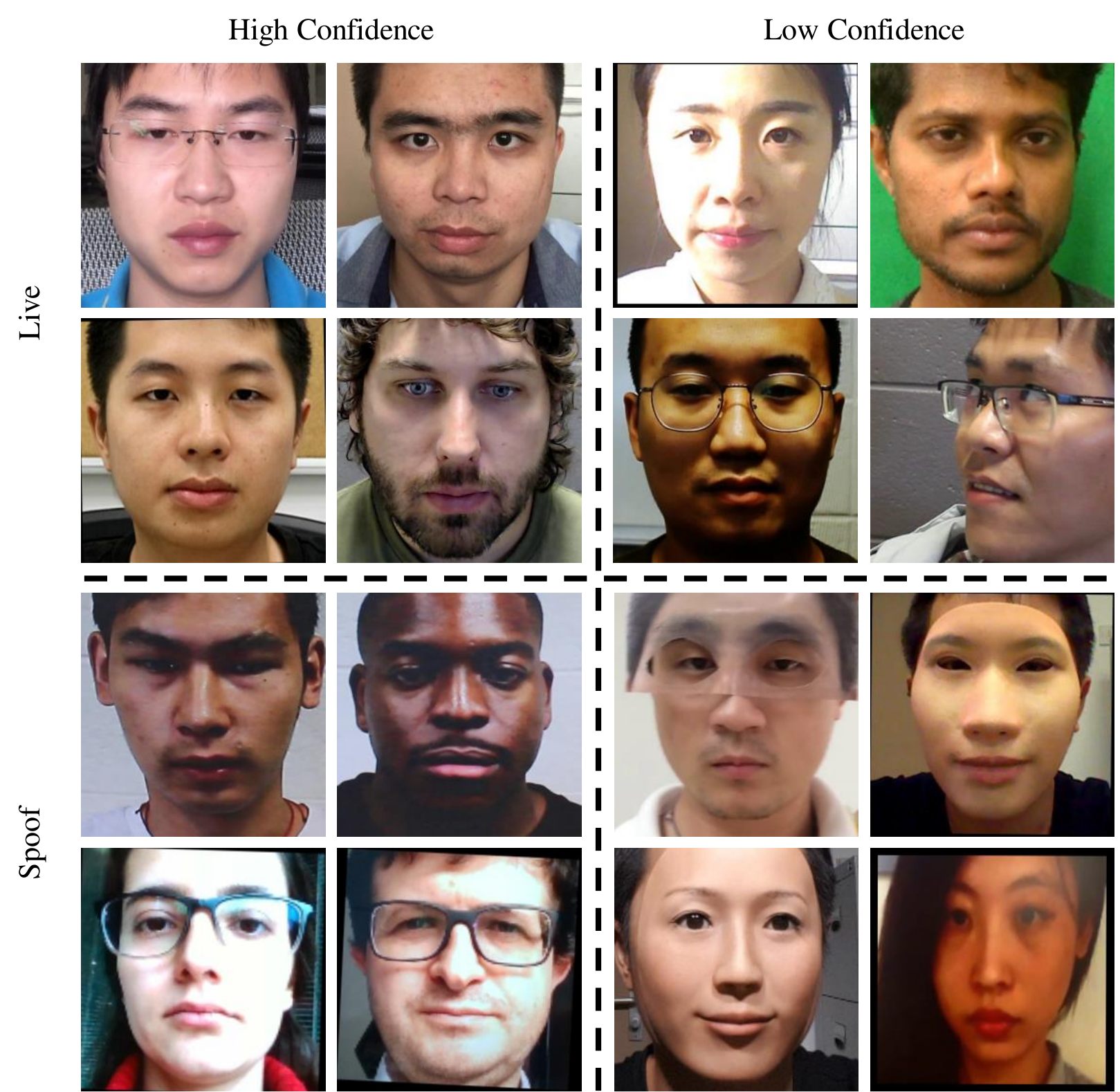}
  \caption{
  Examples of images with high prediction confidence and low prediction confidence. Compared to samples with high confidence, low confidence samples often include strong side illumination, abnormal backgrounds, excessive face tilting, or unknown presentation attacks.}
  \label{fig:Example_Img}
\end{figure}

We randomly select some images with high or low prediction confidence and display them in \Cref{fig:Example_Img}. It can be observed that images with high confidence often exhibit characteristics such as well-aligned faces and uniform illumination. On the other hand, low confidence images tend to have features such as strong side illumination, abnormal backgrounds, and excessive face tilting. Low confidence spoof faces also include some instances of unknown attacks like 3D masks.

\subsection{Ablation Study}

\begin{table*}[t]
\centering
\caption{Comparison of Different Feature Distribution Constraints.}
\resizebox{0.95\hsize}{!}{
\begin{tabular}{cccccccccccc}
        \toprule
\multirow{2}{*}{Method}       & \multicolumn{1}{c}{\multirow{2}{*}{p}} & \multicolumn{2}{c}{\textbf{HKBU-MARs}}                  & \multicolumn{2}{c}{\textbf{ROSE-Youtu}}           & \multicolumn{2}{c}{\textbf{SiW}}                   & \multicolumn{2}{c}{\textbf{WMCA}}                  & \multicolumn{2}{c}{\textbf{Mean}}                  \\
         \cmidrule(r){3-4}\cmidrule(r){5-6}\cmidrule(r){7-8}\cmidrule(r){9-10}\cmidrule(r){11-12}
                              & \multicolumn{1}{c}{}                   & \multicolumn{1}{c}{HTER} & \multicolumn{1}{c}{AUC} & \multicolumn{1}{c}{HTER} & \multicolumn{1}{c}{AUC} & \multicolumn{1}{c}{HTER} & \multicolumn{1}{c}{AUC} & \multicolumn{1}{c}{HTER} & \multicolumn{1}{c}{AUC} & \multicolumn{1}{c}{HTER} & \multicolumn{1}{c}{AUC} \\
         \midrule
\multirow{4}{*}{one-class}    & 0.0                                    & 22.44                    & 85.02                   & 14.27                    & 93.59                   & 8.81                     & 97.15                   & 20.00                    & 87.20                   & 16.38                    & 90.74                   \\
                              & 0.3                                    & 28.72                    & 76.62                   & 23.52                    & 83.92                   & 11.73                    & 95.16                   & 16.04                    & 90.23                   & 20.00                    & 86.48                   \\
                              & 0.5                                    & 32.76                    & 73.85                   & 29.09                    & 77.32                   & 14.96                    & 92.46                   & 13.09                    & 91.92                   & 22.47                    & 83.89                   \\
                              & 0.7                                    & 41.44                    & 64.23                   & 35.16                    & 70.65                   & 20.33                    & 87.45                   & 13.00                    & 92.12                   & 27.48                    & 78.61                   \\
         \midrule
\multirow{4}{*}{domain-based} & 0.0                                    & 23.94                    & 83.74                   & 14.06                    & 93.40                   & 8.23                     & 97.23                   & 23.50                    & 86.18                   & 17.43                    & 90.14                   \\
                              & 0.3                                    & 27.26                    & 82.45                   & 11.32                    & 94.87                   & 5.05                     & 98.66                   & 16.48                    & 92.01                   & 15.02                    & 92.00                   \\
                              & 0.5                                    & 32.33                    & 73.60                   & 9.13                     & 95.76                   & 3.18                     & 99.24                   & 20.14                    & 90.70                   & 16.19                    & 89.82                   \\
                              & 0.7                                    & 31.44                    & 75.28                   & 6.18                     & 96.72                   & \textbf{0.90}            & \textbf{99.90}          & 29.98                    & 80.21                   & 17.13                    & 88.03                   \\
         \midrule
\multirow{4}{*}{binary}       & 0.0                                    & 22.22                    & 86.24                   & 13.00                    & 94.02                   & 7.82                     & 97.81                   & 20.50                    & 88.75                   & 15.88                    & 91.70                   \\
                              & 0.3                                    & 17.39                    & 90.58                   & 12.77                    & 95.00                   & 8.27                     & 97.77                   & 20.58                    & 89.39                   & 14.75                    & 93.18                   \\
                              & 0.5                                    & 11.04                    & 93.60                   & 15.06                    & 93.26                   & 20.55                    & 89.10                   & 17.75                    & 92.64                   & 16.10                    & 92.15                   \\
                              & 0.7                                    & \textbf{5.17}            & \textbf{96.35}          & 25.64                    & 84.14                   & 22.52                    & 82.13                   & \textbf{2.76}            & \textbf{99.47}          & 14.02                    & 90.52                   \\
         \midrule
\multirow{4}{*}{attack-based} & 0.0                                    & 23.82                    & 83.94                   & 14.88                    & 91.37                   & 7.77                     & 97.57                   & 17.96                    & 90.34                   & 16.11                    & 90.81                   \\
                              & 0.3                                    & 18.66                    & 87.95                   & 8.94                     & 94.64                   & 6.43                     & 98.38                   & 16.92                    & 92.68                   & 12.74                    & 93.41                   \\
                              & 0.5                                    & 15.60                    & 90.90                   & 3.91                     & 98.10                   & 2.61                     & 99.38                   & 13.13                    & 94.94                   & 8.81                     & 95.83                   \\
                              & 0.7                                    & 14.52                    & 93.13                   & \textbf{1.37}            & \textbf{99.10}          & 1.12                     & 99.75                   & 4.71                     & 98.35                   & \textbf{5.43}            & \textbf{97.58}          \\
\bottomrule
\end{tabular}
}
\label{tab:ablation}
\end{table*}

We conduct ablation experiments to evaluate the performance of CA-FAS using different approaches to modeling the reliable range, and the results are shown in \Cref{tab:ablation}. \textcolor{black}{Specifically, we divide the training Gaussians from different perspectives, such as categorizing by different domains or distinguishing between live vs. spoof samples, and examine the FAS performance under these conditions.}

\textcolor{black}{First, we try using a single Gaussian distribution to fit all the training data, referred to as ``one-class''.} We find that under this setting, the model's performance on various test datasets is suboptimal. We believe this is because clustering all features together reduces the distinction between live and spoof features, making it difficult for the classifier to differentiate between live and spoof samples.

\textcolor{black}{Next, we test a setting similar to SSDG, where the training spoof samples are divided into different Gaussians based on their domain labels, while the training live samples are modeled using a single Gaussian, referred to as ``domain-based''.} We find that under this setting, the model performs well on SiW, indicating that SSDG's domain-based division method helps with the cross-domain generalization of the features. However, we also observe that the model's performance declines on the other three datasets containing unknown attacks. This suggests that while the domain-based approach aids in handling domain variations, it is less effective when dealing with unknown attack types.

\textcolor{black}{Finally, we cluster the training live samples into a single Gaussian, while the training spoof samples are grouped into another Gaussian. This setting, labeled as ``binary'',} performs well on datasets with unknown mask attacks such as HKBU-MARs and WMCA. However, on the remaining two datasets, while the overall performance on the entire dataset ($p=0.0$) is acceptable, the model's ability to assess its own prediction confidence is poor, leading to a decrease in performance after filtering out low confidence samples.

At the bottom of the table, ``attack-based'' refers to the implementation of the feature constraints employed in our proposed CA-FAS, which demonstrate the most stable performance in the experiments. \textcolor{black}{We believe this is due to the fact that whether an input's pattern resembles any known attack is the key factor in determining the reliability of the model's prediction. Therefore, modeling the reliable range based on known attacks is the most effective method for assessing the model's confidence.}


\subsection{Results on Leave-One-Out (LOO) Setting}

\begin{table*}[t]
\centering
\caption{Comparison Results with the State-of-the-art Methods on the Leave-One-Out Protocol.}
\resizebox{0.85\hsize}{!}{
\begin{tabular}{ccccccccccc}
        \toprule
         & \multicolumn{2}{c}{\textbf{O\&C\&I to M}} & \multicolumn{2}{c}{\textbf{O\&M\&I to C}} & \multicolumn{2}{c}{\textbf{O\&C\&M to I}} & \multicolumn{2}{c}{\textbf{I\&C\&M to O}} & \multicolumn{2}{c}{\textbf{Mean}} \\
         \cmidrule(r){2-3}\cmidrule(r){4-5}\cmidrule(r){6-7}\cmidrule(r){8-9}\cmidrule(r){10-11}
         & HTER               & AUC               & HTER            & AUC             & HTER            & AUC             & HTER            & AUC             & HTER            & AUC             \\
         \midrule
MADDG \cite{MADDG_CVPR2019}    & 17.69          & 88.06           & 24.50           & 84.51            & 22.19            & 84.99            & 27.89           & 80.02           & 23.07                    & 84.40                   \\
SSDG \cite{SSDG_CVPR2020}      & 7.38           & 97.17           & 10.44           & 95.94            & 11.71            & 96.59            & 15.61           & 91.54           & 11.29                    & 95.31                   \\
SSAN \cite{SSAN_CVPR2022}      & 6.67           & 98.75           & 10.00           & 96.67            & 8.88             & 96.79            & 13.72           & 93.63           & 9.82                     & 96.46                   \\
SA-FAS \cite{SAFAS_CVPR2023}   & 5.95           & 96.55           & 8.78            & 95.37            & 6.58             & 97.54            & 10.00           & 96.23           & 7.83                     & 96.42                   \\
GAC-FAS \cite{GACFAS_CVPR2024} & 5.00           & 97.56           & 8.20            & 95.16            & 4.29             & 98.87            & \textbf{8.60}   & \textbf{97.16}  & 6.52                     & 97.19                   \\
         \midrule
OC-SCM \cite{OCSCM_CVPR2024}   & 37.86          & 66.22           & 30.31           & 76.18            & 15.00            & 93.40            & 41.39           & 62.22           & 31.14                    & 74.50                   \\
STDN \cite{STDN_PAMI2022}      & 18.69          & 86.08           & 40.32           & 61.48            & 25.00            & 74.91            & 37.78           & 65.17           & 30.45                    & 71.91                   \\
         \midrule
\textcolor{black}{CA-FAS (p=0.0)}                 & 7.14           & 97.42           & 11.68           & 94.55            & 13.86            & 93.67            & 11.67           & 94.53           & 11.09                    & 95.04                   \\
\textcolor{black}{CA-FAS (p=0.3)}                 & 5.90           & 97.52           & 9.41            & 96.50            & 6.83             & 97.59            & 12.44           & 95.16           & 8.64                     & 96.69                   \\
\textcolor{black}{CA-FAS (p=0.5)}                 & 2.10           & 98.58           & 4.66            & 98.65            & 3.09             & 99.07            & 11.89           & 95.46           & 5.43                     & 97.94                   \\
\textcolor{black}{CA-FAS (p=0.7)}                 & \textbf{1.02}  & \textbf{98.74}  & \textbf{1.06}   & \textbf{99.41}   & \textbf{2.13}    & \textbf{99.49}   & 8.94            & 96.51           & \textbf{3.29}            & \textbf{98.54}          \\
\bottomrule
\end{tabular}
}
\label{tab:LOO}
\end{table*}


Finally, we test the reliability of the proposed CA-FAS under the traditional Leave-One-Out (LOO) setting \cite{LOO_setting2019}, in which the test dataset only includes domain shifts. Mainstream DG-based methods are optimized for this experimental setup, achieving strong domain generalization capabilities and excellent cross-domain performance after training.

The proposed CA-FAS, lacking the domain invariant feature constraints present in other DG-based methods, does not perform as well as the SOTA DG-based methods like SA-FAS and GAC-FAS when tested on the entire test dataset ($p=0.0$). However, despite sacrificing domain generalization capability, CA-FAS gains the ability to perceive domain shifts, which means it can recognize the decline in prediction confidence due to domain shifts. The proposed CA-FAS can thus filter out samples significantly affected by the domain shifts, ensuring the reliability of the model on the remaining samples. When the quantile threshold is set to $p=0.5$, meaning the lowest $50\%$ of test samples in terms of prediction confidence are filtered out, the overall performance of CA-FAS can surpass the SOTA DG-based methods.
The results indicate that under the LOO experimental setup, CA-FAS can still recognize the boundary of its capabilities, achieving reliable liveness detection.


\section{Conclusion}
\label{sec:conclusion}

In this work, we emphasize the importance of reliable face anti-spoofing and address this by proposing a Confidence Aware Face Anti-spoofing (CA-FAS) model, which knows its own capability boundary by measuring the prediction confidence of any input. 
Specifically, we build Gaussian distributions in the feature space to measure the model's reliable range and calculate the prediction confidence based on the Mahalanobis distance between samples and the constructed Gaussians.
We further design experiments that evaluate the performance of FAS models on unseen scenarios and unknown presentation attacks, and the results show that our CA-FAS model achieves much more reliable performance than other FAS models by rejecting samples beyond its reliable range.


\bibliographystyle{IEEEtran}
\bibliography{egbib}


\begin{IEEEbiography}[{\includegraphics[width=1in,height=1.25in,clip,keepaspectratio]{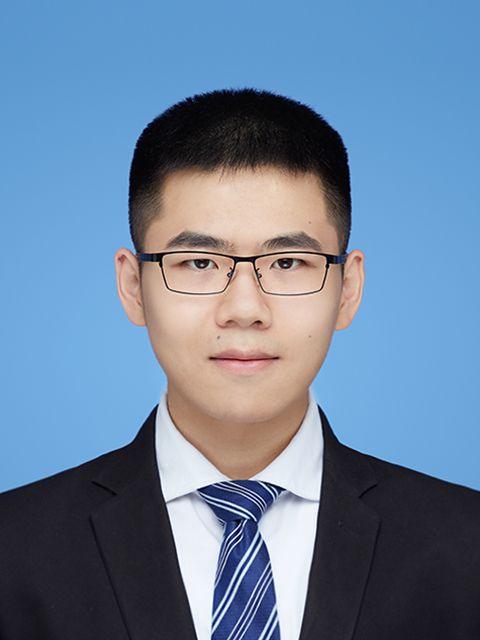}}]{Xingming Long} (Student Member, IEEE) received the B.S. degree in computer science from Tsinghua University in 2021. He is currently pursuing the Ph.D. degree with the Institute of Computing Technology (ICT), Chinese Academy of Sciences (CAS). His research interests include face anti-spoofing and domain generalization.
\end{IEEEbiography}

\begin{IEEEbiography}[{\includegraphics[width=1in,height=1.25in,clip,keepaspectratio]{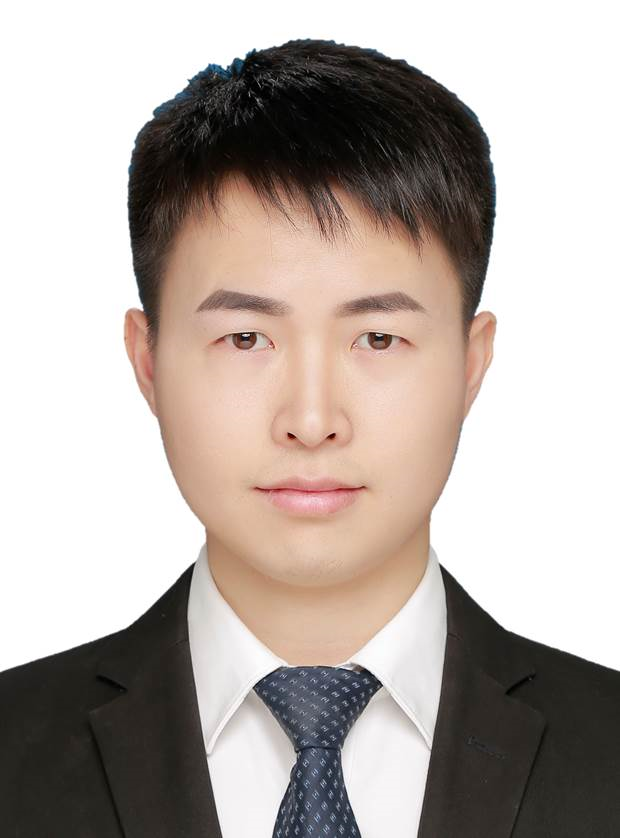}}]{Jie Zhang}
(Member, IEEE) received the Ph.D. degree from the University of Chinese Academy of Sciences, Beijing, China. He is currently an Associate Professor with the Institute of Computing Technology (ICT), Chinese Academy of Sciences (CAS). His research interests include computer vision, pattern recognition, machine learning, particularly include face recognition, image segmentation, weakly/semi-supervised learning, and domain generalization.
\end{IEEEbiography}

\begin{IEEEbiography}[{\includegraphics[width=1in,height=1.25in,clip,keepaspectratio]{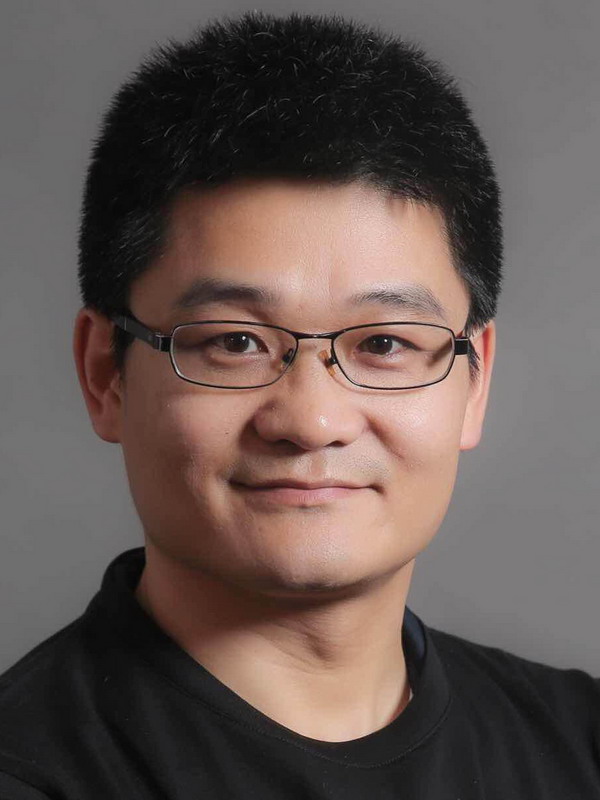}}]{Shiguang Shan}
(Fellow, IEEE) received the Ph.D. degree in computer science from the Institute of Computing Technology (ICT), Chinese Academy of Sciences (CAS), Beijing, China, in 2004. Since 2010, he has been a Full Professor with ICT, CAS, where he is currently the Director of the Key Laboratory of Intelligent Information Processing. His research interests include computer vision, pattern recognition, and machine learning. He has published more than 300 articles in related areas. He was a recipient of the China’s State Natural Science Award in 2015 and the China’s State S\&T Progress Award in 2005 for his research work. He served as an Area Chair for many international conferences, including CVPR, ICCV, AAAI, IJCAI, ACCV, ICPR, and FG. He was/is an Associate Editor of several journals, including IEEE TRANSACTIONS ON IMAGE PROCESSING, Neurocomputing, CVIU, and PRL.
\end{IEEEbiography}

\vfill

\end{document}